\documentclass[runningheads]{llncs}
\usepackage[section]{placeins}
\usepackage{graphicx}
\usepackage{multirow}
\usepackage{amsmath,amssymb} 
\usepackage{color}
\usepackage[width=122mm,left=12mm,paperwidth=146mm,height=193mm,top=12mm,paperheight=217mm]{geometry}
\usepackage{caption}
\usepackage{subcaption}
\usepackage{authblk}

\begin{document}
\pagestyle{headings}
\mainmatter

\title{ShuffleNet V2: Practical Guidelines for Efficient CNN Architecture Design} 

\titlerunning{ }

\authorrunning{ }


\author{Ningning Ma \thanks{Equal contribution.}\inst{1,2}\ \ \ 
        Xiangyu Zhang \inst{\ \star 1}\ \ \ 
        Hai-Tao Zheng \inst{2}\ \ \ 
        Jian Sun \inst{1}}
\institute{ \textsuperscript{1} Megvii Inc (Face++)\hspace{1.5cm}   \textsuperscript{2} Tsinghua University
\\ \email{\{maningning,zhangxiangyu,sunjian\}@megvii.com zheng.haitao@sz.tsinghua.edu.cn}  
 }

\maketitle

\begin{abstract}
Currently, the neural network architecture design is mostly guided by the \emph{indirect} metric of computation complexity, i.e., FLOPs. However, the \emph{direct} metric, e.g., speed, also depends on the other factors such as memory access cost and platform characterics. Thus, this work proposes to evaluate the direct metric on the target platform, beyond only considering FLOPs. Based on a series of controlled experiments, this work derives several practical \emph{guidelines} for efficient network design. Accordingly, a new architecture is presented, called \emph{ShuffleNet V2}. Comprehensive ablation experiments verify that our model is the state-of-the-art in terms of speed and accuracy tradeoff. 
\end{abstract}

\section{Introduction}
\label{sec:intro}

The architecture of deep convolutional neutral networks (CNNs) has evolved for years, becoming more accurate and faster. Since the milestone work of AlexNet~\cite{krizhevsky2012imagenet}, the ImageNet classification accuracy has been significantly improved by novel structures, including VGG~\cite{simonyan2014very}, GoogLeNet~\cite{szegedy2015going}, ResNet~\cite{he2016deep,he2016identity}, DenseNet~\cite{huang2017densely}, ResNeXt~\cite{xie2017aggregated}, SE-Net~\cite{hu2017squeeze}, and automatic neutral architecture search~\cite{zoph2017learning,liu2017progressive,real2018regularized}, to name a few.

Besides accuracy, computation complexity is another important consideration. Real world tasks often aim at obtaining best accuracy under a limited computational budget, given by target platform (e.g., hardware) and application scenarios (e.g., auto driving requires low latency). This motivates a series of works towards light-weight architecture design and better speed-accuracy tradeoff, including Xception~\cite{chollet2016xception}, MobileNet~\cite{howard2017mobilenets}, MobileNet V2~\cite{sandler2018inverted}, ShuffleNet~\cite{zhang2017shufflenet}, and CondenseNet~\cite{huang2017condensenet}, to name a few. Group convolution and depth-wise convolution are crucial in these works.

\begin{figure}[t]
	\centering
	\includegraphics[height=6.9cm]{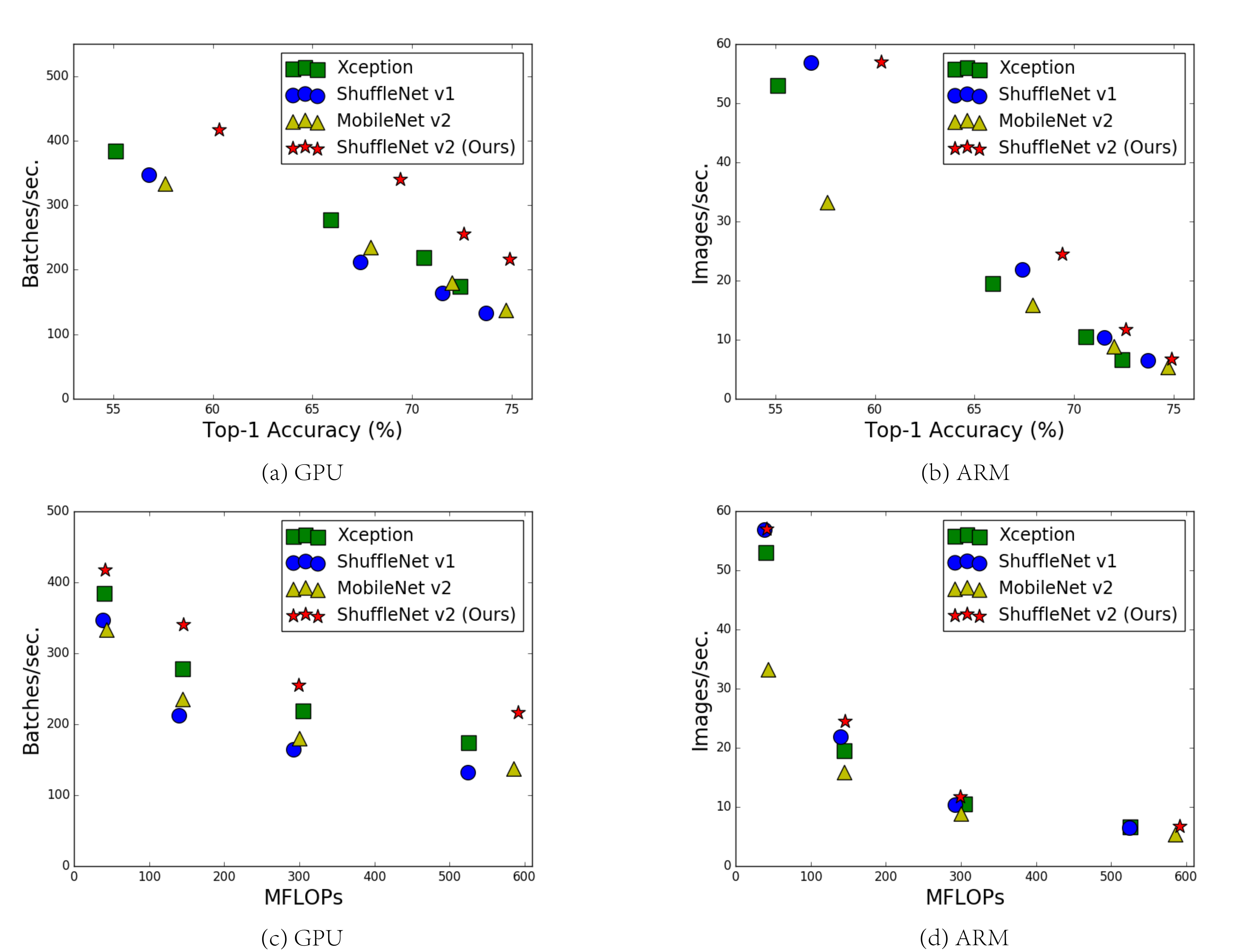}
	\caption{Measurement of accuracy (ImageNet classification on validation set), speed and FLOPs of four network architectures on two hardware platforms with four different level of computation complexities (see text for details).  (a, c) GPU results, $batch size=8$. (b, d) ARM results, $batch size=1$. The best performing algorithm, our proposed ShuffleNet v2, is on the top right region, under all cases.}
	\label{fig:speed}
\end{figure}

To measure the computation complexity, a widely used metric is the number of float-point operations, or \emph{FLOPs}\footnote{In this paper, the definition of \emph{FLOPs} follows~\cite{zhang2017shufflenet}, i.e. the number of multiply-adds.}. However, FLOPs is an \emph{indirect} metric. It is an approximation of, but usually not equivalent to the \emph{direct} metric that we really care about, such as speed or latency. Such discrepancy has been noticed in previous works~\cite{liu2017learning,wen2016learning,sandler2018inverted,he2017channel}. For example, \emph{MobileNet v2}~\cite{sandler2018inverted} is much faster than \emph{NASNET-A}~\cite{zoph2017learning} but they have comparable FLOPs. This phenomenon is further exmplified in Figure~\ref{fig:speed}(c)(d), which show that networks with similar FLOPs have different speeds. Therefore,
using FLOPs as the only metric for computation complexity is insufficient and could lead to sub-optimal design.

The discrepancy between the indirect (FLOPs) and direct (speed) metrics can be attributed to two main reasons. First, several important factors that have considerable affection on speed are not taken into account by FLOPs. One such factor is \emph{memory access cost} (MAC). Such cost constitutes a large portion of runtime in certain operations like group convolution. It could be bottleneck on devices with strong computing power, e.g., GPUs. This cost should not be simply ignored during network architecture design. Another one is \emph{degree of parallelism}. A model with high degree of parallelism could be much faster than another one with low degree of parallelism, under the same FLOPs.

Second, operations with the same FLOPs could have different running time, depending on the platform. For example, tensor decomposition is widely used in early works~\cite{jaderberg2014speeding,zhang2015efficient,zhang2016accelerating} to accelerate the matrix multiplication. However, the recent work~\cite{he2017channel} finds that the decomposition in \cite{zhang2016accelerating} is even slower on GPU although it reduces FLOPs by 75\%. We investigated this issue and found that this is because the latest CUDNN~\cite{chetlur2014cudnn} library is specially optimized for $3\times3$ conv. We cannot certainly think that $3\times3$ conv is 9 times slower than $1\times1$ conv.

With these observations, we propose that two principles should be considered for effective network architecture design. First, the direct metric (e.g., speed) should be used instead of the indirect ones (e.g., FLOPs). Second, such metric should be evaluated on the target platform.

In this work, we follow the two principles and propose a more effective network architecture. In Section~\ref{sec:guidelines}, we firstly analyze the runtime performance of two representative state-of-the-art networks \cite{zhang2017shufflenet,sandler2018inverted}. Then, we derive four guidelines for efficient network design, which are beyond only considering FLOPs. While these guidelines are platform independent, we perform a series of controlled experiments to validate them on two different platforms (GPU and ARM) with dedicated code optimization, ensuring that our conclusions are state-of-the-art.

In Section~\ref{sec:shufflenet}, according to the guidelines, we design a new network structure. As it is inspired by ShuffleNet~\cite{zhang2017shufflenet}, it is called \emph{ShuffleNet V2}. It is demonstrated much faster and more accurate than the previous networks on both platforms, via comprehensive validation experiments in Section~\ref{sec:exp}. Figure~\ref{fig:speed}(a)(b) gives an overview of comparison. For example, given the computation complexity budget of 40M FLOPs, ShuffleNet v2 is 3.5\% and 3.7\% more accurate than ShuffleNet v1 and MobileNet v2, respectively.


\begin{figure}[t]
	\centering
	\includegraphics[height=3.5cm]{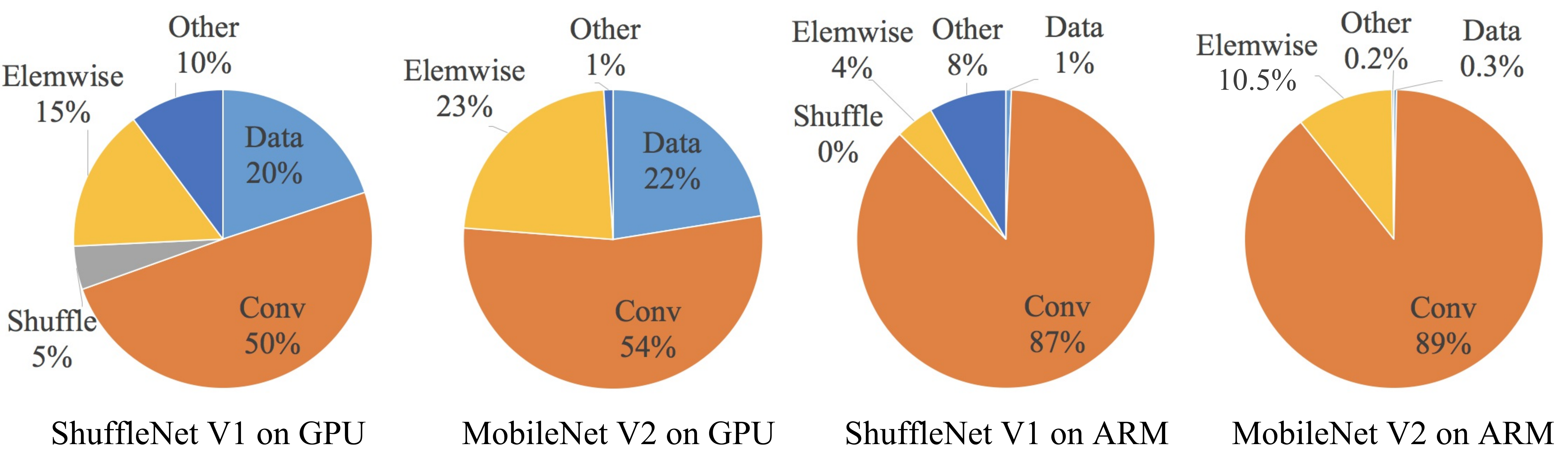}
	\caption{Run time decomposition on two representative state-of-the-art network architectures, \emph{ShuffeNet v1}~\cite{zhang2017shufflenet} (1$\times$, $g=3$) and \emph{MobileNet v2}~\cite{sandler2018inverted} (1$\times$).}
	\label{fig:opr_occupy}
\end{figure}

\section{Practical Guidelines for Efficient Network Design}
\label{sec:guidelines}

Our study is performed on two widely adopted hardwares with industry-level optimization of CNN library. We note that our CNN library is more efficient than most open source libraries. Thus, we ensure that our observations and conclusions are solid and of significance for practice in industry.

\begin{itemize}
\item \emph{GPU.} A single NVIDIA GeForce GTX 1080Ti is used. The convolution library is CUDNN 7.0 \cite{chetlur2014cudnn}. We also activate the benchmarking function of CUDNN to select the fastest algorithms for different convolutions respectively. 

\item \emph{ARM.} A Qualcomm Snapdragon 810. We use a highly-optimized Neon-based implementation. A single thread is used for evaluation.
\end{itemize}

Other settings include: full optimization options (e.g. tensor fusion, which is used to reduce the overhead of small operations) are switched on. The input image size is $224\times 224$. Each network is randomly initialized and evaluated for 100 times. The average runtime is used.

To initiate our study, we analyze the runtime performance of two state-of-the-art networks, \emph{ShuffleNet v1}~\cite{zhang2017shufflenet} and \emph{MobileNet v2}~\cite{sandler2018inverted}. They are both highly efficient and accurate on ImageNet classification task. They are both widely used on low end devices such as mobiles. Although we only analyze these two networks, we note that they are representative for the current trend. At their core are group convolution and depth-wise convolution, which are also crucial components for other state-of-the-art networks, such as ResNeXt~\cite{xie2017aggregated}, Xception~\cite{chollet2016xception}, MobileNet~\cite{howard2017mobilenets}, and CondenseNet~\cite{huang2017condensenet}.

The overall runtime is decomposed for different operations, as shown in Figure~\ref{fig:opr_occupy}. We note that the FLOPs metric only account for the convolution part. Although this part consumes most time, the other operations including data I/O, data shuffle and element-wise operations (AddTensor, ReLU, etc) also occupy considerable amount of time. Therefore, FLOPs is not an accurate enough estimation of actual runtime.

Based on this observation, we perform a detailed analysis of runtime (or speed) from several different aspects and derive several practical guidelines for efficient network architecture design.

\begin{table}[t]
\centering
\begin{tabular}{c|c|ccc|c|ccc}
\hline
         &    & \multicolumn{3}{c|}{GPU (Batches/sec.)} &  & \multicolumn{3}{|c}{ARM (Images/sec.)} \\ \hline
   c1:c2   &  (c1,c2) for $\times 1$ & $\times1$       & $\times2$       & $\times4$  & (c1,c2) for $\times 1$  & $\times1$   & $\times2$   & $\times4$   \\ \hline
 1:1 & (128,128)   & 1480        & 723         & 232  &  (32,32)    & 76.2        & 21.7       & 5.3        \\
 1:2 &  (90,180)  & 1296        & 586         & 206   &  (22,44)   & 72.9        & 20.5       & 5.1        \\ 
 1:6 & (52,312)    & 876         & 489         & 189   &  (13,78)   & 69.1        & 17.9       & 4.6        \\  
 1:12  &  (36,432) & 748         & 392         & 163   &  (9,108)   & 57.6        & 15.1       & 4.4        \\ \hline
\end{tabular}

\caption{Validation experiment for \textbf{Guideline 1}. Four different ratios of number of input/output channels ($c1$ and $c2$) are tested, while the total FLOPs under the four ratios is fixed by varying the number of channels. Input image size is $56\times56$.}
\label{table:guideline_channel_width}
\end{table}

\textbf{G1) Equal channel width minimizes memory access cost (MAC).}
The modern networks usually adopt \emph{depthwise separable convolutions} ~\cite{chollet2016xception,howard2017mobilenets,zhang2017shufflenet,sandler2018inverted}, where the pointwise convolution (i.e., $1\times1$ convolution) accounts for most of the complexity~\cite{zhang2017shufflenet}. We study the kernel shape of the $1\times1$ convolution. The shape is specified by two parameters: the number of input channels $c_1$ and output channels $c_2$. Let $h$ and $w$ be the spatial size of the feature map, the FLOPs of the $1\times1$ convolution is $B=hwc_1c_2$.

For simplicity, we assume the cache in the computing device is large enough to store the entire feature maps and parameters. Thus, the memory access cost (MAC), or the number of memory access operations, is $\text{MAC} = hw(c_1 + c_2) + c_1c_2$. Note that the two terms correspond to the memory access for input/output feature maps and kernel weights, respectively.

From mean value inequality, we have

\begin{equation}
\label{equ:convshape}
\text{MAC} \ge 2\sqrt{hwB} + \frac{B}{hw}.
\end{equation}

Therefore, \emph{MAC has a lower bound given by FLOPs. It reaches the lower bound when the numbers of input and output channels are equal.}

The conclusion is theoretical. In practice, the cache on many devices is not large enough. Also, modern computation libraries usually adopt complex blocking strategies to make full use of the cache mechanism \cite{das2016distributed}. Therefore, the real MAC may deviate from the theoretical one. To validate the above conclusion, an experiment is performed as follows. A benchmark network is built by stacking 10 building blocks repeatedly. Each block contains two convolution layers. The first contains $c_1$ input channels and $c_2$ output channels, and the second otherwise. 

Table~\ref{table:guideline_channel_width} reports the running speed by varying the ratio $c_1:c_2$ while fixing the total FLOPs. It is clear that when $c_1:c_2$ is approaching $1:1$, the MAC becomes smaller and the network evaluation speed is faster.

\begin{table}[t]
\centering
\begin{tabular}{c|c|ccc|c|ccc}
\hline
       &  & \multicolumn{3}{c|}{GPU (Batches/sec.)} & & \multicolumn{3}{|c}{CPU (Images/sec.)} \\ \hline
 g & c for $\times1$ & $\times1$       & $\times2$       & $\times4$      & c for $\times1$ & $\times1$        & $\times2$      & $\times4$      \\ \hline
  1 & 128 & 2451        & 1289        & 437   &   64  & 40.0        & 10.2       & 2.3        \\
  2 & 180 & 1725        & 873         & 341    &  90  & 35.0        & 9.5        & 2.2        \\ 
  4 & 256 & 1026        & 644         & 338    &  128  & 32.9        & 8.7        & 2.1        \\ 
  8 & 360 & 634         & 445         & 230    &  180  & 27.8        & 7.5        & 1.8        \\ \hline
\end{tabular}
\caption{Validation experiment for \textbf{Guideline 2}. Four values of group number $g$ are tested, while the total FLOPs under the four values is fixed by varying the total channel number $c$. Input image size is $56\times56$.}
\label{table:guideline_group_number}
\end{table}

\textbf{G2) Excessive group convolution increases MAC.} 
Group convolution is at the core of modern network architectures~\cite{xie2017aggregated,zhang2017shufflenet,ioannou2016deep,zhang2017interleaved,xie2018igcv,sun2018igcv3}. It reduces the computational complexity (FLOPs) by changing the dense convolution between all channels to be sparse (only within groups of channels). On one hand, it allows usage of more channels given a fixed FLOPs and increases the network capacity (thus better accuracy). On the other hand, however, the increased number of channels results in more MAC. 

Formally, following the notations in \textbf{G1} and Eq.~\ref{equ:convshape}, the relation between MAC and FLOPs for $1\times 1$ group convolution is

\begin{equation}
\begin{aligned}
\text{MAC} &= hw(c_1 + c_2) + \frac{c_1 c_2}{g} \\
&= hwc_1 + \frac{Bg}{c_1}  + \frac{B}{hw},
\end{aligned}
\end{equation}
where $g$ is the number of groups and $B=hwc_1c_2/g$ is the FLOPs. It is easy to see that, given the fixed input shape $c_1 \times h \times w$ and the computational cost $B$, MAC increases with the growth of $g$. 

To study the affection in practice, a benchmark network is built by stacking 10 pointwise group convolution layers. Table~\ref{table:guideline_group_number} reports the running speed of using different group numbers while fixing the total FLOPs. It is clear that using a large group number decreases running speed significantly. For example, using 8 groups is more than two times slower than using 1 group (standard dense convolution) on GPU and up to 30\% slower on ARM. This is mostly due to increased MAC. We note that our implementation has been specially optimized and is much faster than trivially computing convolutions group by group.


Therefore, we suggest that \emph{the group number should be carefully chosen based on the target platform and task. It is unwise to use a large group number simply because this may enable using more channels, because the benefit of accuracy increase can easily be outweighed by the rapidly increasing computational cost.}

\begin{table}[t]
\centering
\begin{tabular}{l|ccc|ccc}
\hline
                                           & \multicolumn{3}{c}{GPU (Batches/sec.)} & \multicolumn{3}{|c}{CPU (Images/sec.)} \\ \hline
            & c=128       & c=256       & c=512      & c=64        & c=128      & c=256      \\ \hline
  1-fragment          & 2446        & 1274        & 434        & 40.2        & 10.1       & 2.3        \\ 
  2-fragment-series   & 1790        & 909         & 336        & 38.6        & 10.1       & 2.2        \\ 
  4-fragment-series   & 752         & 745         & 349        & 38.4        & 10.1       & 2.3        \\ 
  2-fragment-parallel & 1537        & 803         & 320        & 33.4        & 9.1        & 2.2        \\ 
  4-fragment-parallel & 691         & 572         & 292        & 35.0        & 8.4        & 2.1        \\ \hline
\end{tabular}
\caption{Validation experiment for \textbf{Guideline 3}. $c$ denotes the number of channels for \emph{1-fragment}. The channel number in other fragmented structures is adjusted so that the FLOPs is the same as \emph{1-fragment}. Input image size is $56\times56$.}
\label{table:guideline_fragmentation}
\end{table}

\begin{table}[t]
\centering
\begin{tabular}{c|c|ccc|ccc}
\hline
 &  & \multicolumn{3}{c}{GPU (Batches/sec.)} & \multicolumn{3}{|c}{CPU (Images/sec.)} \\ \hline
ReLU & short-cut & c=32        & c=64       & c=128       & c=32        & c=64       & c=128      \\ \hline
yes & yes & 2427        & 2066       & 1436        & 56.7        & 16.9       & 5.0        \\ \hline 
yes   & no   & 2647        & 2256       & 1735        & 61.9        & 18.8       & 5.2        \\ \hline 
no & yes  & 2672        & 2121       & 1458        & 57.3        & 18.2       & 5.1        \\ \hline 
no & no   & 2842        & 2376       & 1782        & 66.3        & 20.2       & 5.4        \\ \hline
\end{tabular}
\caption{Validation experiment for \textbf{Guideline 4}. The ReLU and shortcut operations are removed from the ``bottleneck'' unit~\cite{he2016deep}, separately. $c$ is the number of channels in unit. The unit is stacked repeatedly for 10 times to benchmark the speed.}
\label{table:guideline_elementwise}
\end{table}

\textbf{G3) Network fragmentation reduces degree of parallelism.} In the  GoogLeNet series~\cite{szegedy2017inception,szegedy2016rethinking,szegedy2015going,ioffe2015batch} and auto-generated architectures~\cite{zoph2017learning,real2018regularized,liu2017progressive}), a ``multi-path" structure is widely adopted in each network block. A lot of small operators (called ``fragmented operators'' here) are used instead of a few large ones. For example, in \emph{NASNET-A} \cite{zoph2017learning} the number of fragmented operators (i.e. the number of individual convolution or pooling operations in one building block) is 13. In contrast, in regular structures like ResNet~\cite{he2016deep}, this number is 2 or 3.

Though such fragmented structure has been shown beneficial for accuracy, it could decrease efficiency because it is unfriendly for devices with strong parallel computing powers like GPU. It also introduces extra overheads such as kernel launching and synchronization.

To quantify how network fragmentation affects efficiency, we evaluate a series of network blocks with different degrees of fragmentation. Specifically, each building block consists of from $1$ to $4$ $1\times1$ convolutions, which are arranged in sequence or in parallel. The block structures are illustrated in appendix. Each block is repeatedly stacked for 10 times. Results in Table~\ref{table:guideline_fragmentation} show that fragmentation reduces the speed significantly on GPU, e.g. 4-fragment structure is 3$\times$ slower than 1-fragment. On ARM, the speed reduction is relatively small.

\begin{figure}[t]
	\centering
	\includegraphics[height=5.5cm]{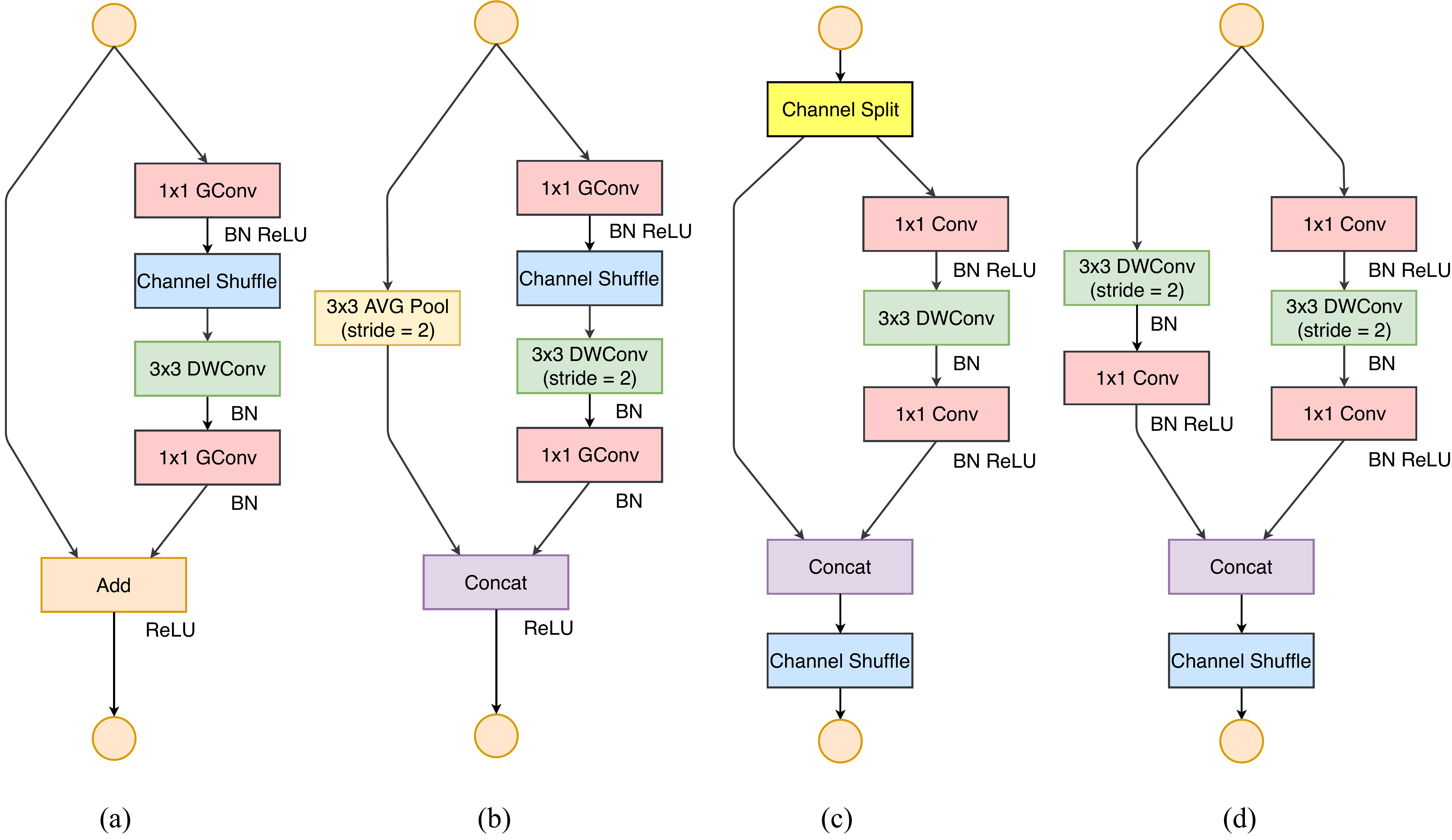}
	\caption{Building blocks of ShuffleNet v1~\cite{zhang2017shufflenet} and this work. (a): the basic ShuffleNet unit; (b) the ShuffleNet unit for spatial down sampling ($2\times$); (c) our basic unit; (d) our unit for spatial down sampling ($2\times$). \textbf{DWConv}: depthwise convolution. \textbf{GConv}: group convolution. } 
\label{fig:shufflenet}
\end{figure}

\textbf{G4) Element-wise operations are non-negligible.}
As shown in Figure~\ref{fig:opr_occupy}, in light-weight models like~\cite{zhang2017shufflenet,sandler2018inverted}, element-wise operations occupy considerable amount of time, especially on GPU. Here, the element-wise operators include ReLU, AddTensor, AddBias, etc. They have small FLOPs but relatively heavy MAC. Specially, we also consider \emph{depthwise convolution}~\cite{chollet2016xception,howard2017mobilenets,sandler2018inverted,zhang2017shufflenet} as an element-wise operator as it also has a high MAC/FLOPs ratio.

For validation, we experimented with the ``bottleneck'' unit ($1\times1$ conv followed by $3\times3$ conv followed by $1\times1$ conv, with ReLU and shortcut connection) in ResNet~\cite{he2016deep}. The ReLU and shortcut operations are removed, separately. Runtime of different variants is reported in Table~\ref{table:guideline_elementwise}. We observe around 20\% speedup is obtained on both GPU and ARM, after ReLU and shortcut are removed.

\textbf{Conclusion and Discussions} Based on the above guidelines and empirical studies, we conclude that an efficient network architecture should 1) use ''balanced`` convolutions (equal channel width); 2) be aware of the cost of using group convolution; 3) reduce the degree of fragmentation; and 4) reduce element-wise operations. These desirable properties depend on platform characterics (such as memory manipulation and code optimization) that are beyond theoretical FLOPs. They should be taken into accout for practical network design.

Recent advances in light-weight neural network architectures~\cite{zhang2017shufflenet,howard2017mobilenets,sandler2018inverted,zoph2017learning,real2018regularized,liu2017progressive,chollet2016xception} are mostly based on the metric of FLOPs and do not consider these properties above. For example, \emph{ShuffleNet v1}~\cite{zhang2017shufflenet} heavily depends group convolutions (against \textbf{G2}) and bottleneck-like building blocks (against \textbf{G1}). \emph{MobileNet v2}~\cite{sandler2018inverted} uses an  inverted bottleneck structure that violates \textbf{G1}. It uses depthwise convolutions and ReLUs on ``thick" feature maps. This violates \textbf{G4}. The auto-generated structures ~\cite{zoph2017learning,real2018regularized,liu2017progressive} are highly fragmented and violate \textbf{G3}.


\begin{table}[t]
\centering
\begin{tabular}{l|c|c|c|c|c|c|c|c}
\hline
\multirow{2}{*}{Layer} & \multirow{2}{*}{Output size}& \multirow{2}{*}{KSize} & \multirow{2}{*}{Stride} & \multirow{2}{*}{Repeat} & \multicolumn{4}{c}{Output channels} \\ 
\cline{6-9} & & & & & 0.5$\times$    & 1$\times$     & 1.5$\times$     & 2$\times$     \\ 
\hline
Image & 224$\times$224 & & & & 3 & 3 & 3 & 3 \\ \hline
\begin{tabular}[l]{@{}l@{}}Conv1\\ MaxPool\end{tabular} & \begin{tabular}[c]{@{}c@{}}112$\times$112\\ 56$\times$56\end{tabular} & \begin{tabular}[c]{@{}c@{}}3$\times$3\\ 
3$\times$3\end{tabular} & \begin{tabular}[c]{@{}c@{}}2\\ 2\end{tabular} & 1 & 24 & 24 & 24 & 24 \\ \hline
Stage2 & \begin{tabular}[c]{@{}c@{}}28$\times$28\\ 28$\times$28\end{tabular} & & \begin{tabular}[c]{@{}c@{}}2\\ 1\end{tabular} & \begin{tabular}[c]{@{}c@{}}1\\ 
3\end{tabular} & 48      & 116    & 176      & 244    \\ 
\hline
Stage3 & \begin{tabular}[c]{@{}c@{}}14$\times$14\\ 14$\times$14\end{tabular}   & & \begin{tabular}[c]{@{}c@{}}2\\ 1\end{tabular} & \begin{tabular}[c]{@{}c@{}}1\\ 7\end{tabular} & 96 & 232 & 352      & 488 \\
\hline
Stage4 & \begin{tabular}[c]{@{}c@{}}7$\times$7\\ 7$\times$7\end{tabular} & & \begin{tabular}[c]{@{}c@{}}2\\ 1\end{tabular} & \begin{tabular}[c]{@{}c@{}}1\\ 3\end{tabular} & 192 & 464 & 704 & 976   \\ 
\hline
Conv5 & 7$\times$7 & 1$\times$1 & 1& 1 & 1024    & 1024   & 1024     & 2048   \\ 
\hline
GlobalPool & 1$\times$1 & 7$\times$7 & & & & & & \\ 
\hline
FC & & & & & 1000 & 1000   & 1000 & 1000   \\ 
\hline
FLOPs & & & & &   41M      &  146M      & 299M   & 591M \\ 
\hline
\# of Weights & & & & &   1.4M      &  2.3M      & 3.5M   & 7.4M \\ 
\hline
\end{tabular}
\caption{Overall architecture of ShuffleNet v2, for four different levels of complexities.}
\label{tbl:arch}
\end{table}

\section{ShuffleNet V2: an Efficient Architecture}
\label{sec:shufflenet}

\textbf{Review of ShuffleNet v1}~\cite{zhang2017shufflenet}. ShuffleNet is a state-of-the-art network architecture. It is widely adopted in low end devices such as mobiles. It inspires our work. Thus, it is reviewed and analyzed at first.

According to~\cite{zhang2017shufflenet}, the main challenge for light-weight networks is that only a limited number of feature channels is affordable under a given computation budget (FLOPs). To increase the number of channels without significantly increasing FLOPs, two techniques are adopted in~\cite{zhang2017shufflenet}: pointwise group convolutions and bottleneck-like structures. A ``channel shuffle'' operation is then introduced to enable information communication between different groups of channels and improve accuracy. The building blocks are illustrated in Figure~\ref{fig:shufflenet}(a)(b).

As discussed in Section~\ref{sec:guidelines}, both pointwise group convolutions and bottleneck structures increase MAC (\textbf{G1} and \textbf{G2}). This cost is non-negligible, especially for light-weight models. Also, using too many groups violates \textbf{G3}. The element-wise ``Add'' operation in the shortcut connection is also undesirable (\textbf{G4}). Therefore, in order to achieve high model capacity and efficiency, the key issue is how to maintain a large number and equally wide channels with neither dense convolution nor too many groups.

\textbf{Channel Split and ShuffleNet V2} Towards above purpose, we introduce a simple operator called \emph{channel split}. It is illustrated in Figure~\ref{fig:shufflenet}(c). At the beginning of each unit, the input of $c$ feature channels are split into two branches with $c - c'$ and $c'$ channels, respectively. Following \textbf{G3}, one branch remains as identity. The other branch consists of three convolutions with the same input and output channels to satisfy \textbf{G1}. The two $1\times1$ convolutions are no longer group-wise, unlike~\cite{zhang2017shufflenet}. This is partially to follow \textbf{G2}, and partially because the split operation already produces two groups.

After convolution, the two branches are concatenated. So, the number of channels keeps the same (\textbf{G1}). The same ``channel shuffle'' operation as in~\cite{zhang2017shufflenet} is then used to enable information communication between the two branches.

After the shuffling, the next unit begins. Note that the ``Add'' operation in ShuffleNet v1~\cite{zhang2017shufflenet} no longer exists. Element-wise operations like ReLU and \emph{depthwise convolutions} exist only in one branch. Also, the three successive element-wise operations, ``Concat'', ``Channel Shuffle'' and ``Channel Split'', are merged into a single element-wise operation. These changes are beneficial according to \textbf{G4}.

For spatial down sampling, the unit is slightly modified and illustrated in Figure~\ref{fig:shufflenet}(d). The channel split operator is removed. Thus, the number of output channels is doubled.

The proposed building blocks (c)(d), as well as the resulting networks, are called \emph{ShuffleNet V2}. Based the above analysis, we conclude that this architecture design is highly efficient as it follows all the guidelines.

The building blocks are repeatedly stacked to construct the whole network. For simplicity, we set $c'=c/2$. The overall network structure is similar to ShuffleNet v1~\cite{zhang2017shufflenet} and summarized in Table~\ref{tbl:arch}. There is only one difference: an additional $1\times 1$ convolution layer is added right before \emph{global averaged pooling} to mix up features, which is absent in ShuffleNet v1. Similar to \cite{zhang2017shufflenet}, the number of channels in each block is scaled to generate networks of different complexities, marked as 0.5$\times$, 1$\times$, etc.

\begin{figure}[t]
	\centering
	\includegraphics[height=4.5cm]{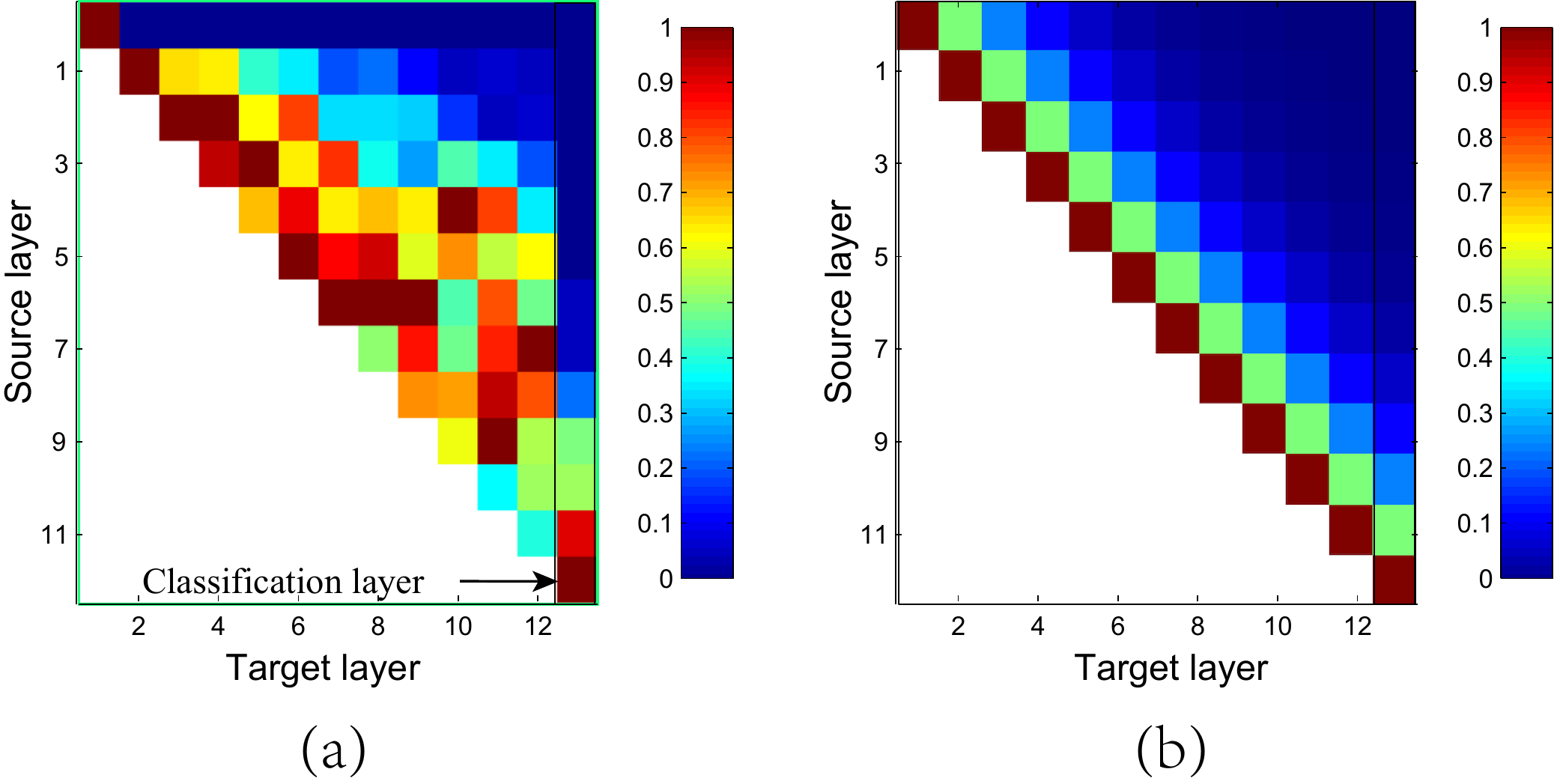}
	\caption{Illustration of the patterns in feature reuse for \emph{DenseNet}~\cite{huang2017densely} and ShuffleNet V2. (a) (courtesy of~\cite{huang2017densely}) the average absolute filter weight of convolutional layers in a model. The color of pixel $(s, l)$ encodes the average $l1$-norm of weights connecting layer $s$ to $l$. (b) The color of pixel $(s, l)$ means the number of channels \emph{directly} connecting block $s$ to block $l$ in ShuffleNet v2. All pixel values are normalized to $[0,1]$.  } 
	\label{fig:densenet}
\end{figure}

\textbf{Analysis of Network Accuracy} ShuffleNet v2 is not only efficient, but also accurate. There are two main reasons. First, the high efficiency in each building block enables using more feature channels and larger network capacity.

Second, in each block, half of feature channels (when $c'=c/2$) directly go through the block and join the next block. This can be regarded as a kind of \emph{feature reuse}, in a similar spirit as in \emph{DenseNet}~\cite{huang2017densely} and \emph{CondenseNet}~\cite{huang2017condensenet}.

In \emph{DenseNet}\cite{huang2017densely}, to analyze the feature reuse pattern, the $l1$-norm of the weights between layers are plotted, as in Figure~\ref{fig:densenet}(a). It is clear that the connections between the adjacent layers are stronger than the others. This implies that the dense connection between all layers could introduce redundancy. The recent \emph{CondenseNet}~\cite{huang2017condensenet} also supports the viewpoint. 

In ShuffleNet V2, it is easy to prove that the number of ``directly-connected'' channels between $i$-th and $(i+j)$-th building block is $r^jc$, where $r=(1-c')/c$. In other words, the amount of feature reuse decays exponentially with the distance between two blocks. Between distant blocks, the feature reuse becomes much weaker. Figure~\ref{fig:densenet}(b) plots the similar visualization as in (a), for $r=0.5$. Note that the pattern in (b) is similar to (a).

Thus, the structure of ShuffleNet V2 realizes this type of feature re-use pattern \emph{by design}. It shares the similar benefit of feature re-use for high accuracy as in \emph{DenseNet}~\cite{huang2017densely}, but it is much more efficient as analyzed earlier. This is verified in experiments, Table \ref{tbl:cls_comp}.

\section{Experiment}
\label{sec:exp}

Our ablation experiments are performed on ImageNet 2012 classification dataset~\cite{deng2009imagenet,russakovsky2015imagenet}. Following the common practice~\cite{zhang2017shufflenet,howard2017mobilenets,sandler2018inverted}, all networks in comparison have four levels of computational complexity, i.e. about 40, 140, 300 and 500+ MFLOPs. Such complexity is typical for mobile scenarios. Other hyper-parameters and protocols are exactly the same as \emph{ShuffleNet v1} \cite{zhang2017shufflenet}.

We compare with following network architectures~\cite{chollet2016xception,sandler2018inverted,huang2017densely,zhang2017shufflenet}:

\begin{itemize}
\item \emph{ShuffleNet v1} \cite{zhang2017shufflenet}. In \cite{zhang2017shufflenet}, a series of group numbers $g$ is compared. It is suggested that the $g=3$ has better trade-off between accuracy and speed. This also agrees with our observation. In this work we mainly use $g=3$.

\item \emph{MobileNet v2} \cite{sandler2018inverted}. It is better than \emph{MobileNet v1} \cite{howard2017mobilenets}. For comprehensive comparison, we report accuracy in both original paper~\cite{sandler2018inverted} and our reimplemention, as some results in~\cite{sandler2018inverted} are not available.

\item \emph{Xception} \cite{chollet2016xception}. The original \emph{Xception} model \cite{chollet2016xception} is very large (FLOPs $>$2G), which is out of our range of comparison. The recent work~\cite{li2017light} proposes a modified light weight Xception structure that shows better trade-offs between accuracy and efficiency. So, we compare with this variant.

\item \emph{DenseNet} \cite{huang2017densely}. The original work~\cite{huang2017densely} only reports results of large models (FLOPs $>$2G). For direct comparison, we reimplement it following the architecture settings in Table \ref{tbl:arch}, where the building blocks in Stage 2-4 consist of \emph{DenseNet} blocks. We adjust the number of channels to meet different target complexities.
\end{itemize}

Table~\ref{tbl:cls_comp} summarizes all the results. We analyze these results from different aspects.

\paragraph{Accuracy vs. FLOPs.}
It is clear that the proposed ShuffleNet v2 models outperform all other networks by a large margin\footnote{As reported in \cite{sandler2018inverted}, MobileNet v2 of 500+ MFLOPs has comparable accuracy with the counterpart ShuffleNet v2 (25.3\% vs. 25.1\% top-1 error); however, our reimplemented version is not as good (26.7\% error, see Table~\ref{tbl:cls_comp}).}, especially under smaller computational budgets. 
Also, we note that MobileNet v2 performs pooly at 40 MFLOPs level with $224\times 224$ image size. This is probably caused by too few channels. In contrast, our model do not suffer from this drawback as our efficient design allows using more channels. 
Also, while both of our model and DenseNet \cite{huang2017densely} reuse features, our model is much more efficient, as discussed in Sec.~\ref{sec:shufflenet}.

Table~\ref{tbl:cls_comp} also compares our model with other state-of-the-art networks including \emph{CondenseNet}~\cite{huang2017condensenet}, \emph{IGCV2}~\cite{xie2018igcv}, and \emph{IGCV3}~\cite{sun2018igcv3} where appropriate. Our model performs better consistently at various complexity levels. 

\paragraph{Inference Speed vs. FLOPs/Accuracy. }
For four architectures with good accuracy, ShuffleNet v2, MobileNet v2, ShuffleNet v1 and Xception, we compare their actual speed vs. FLOPs, as shown in Figure~\ref{fig:speed}(c)(d). More results on different resolutions are provided in Appendix Table 1.

ShuffleNet v2 is clearly faster than the other three networks, especially on GPU. For example, at 500MFLOPs ShuffleNet v2 is 58\% faster than MobileNet v2, 63\% faster than ShuffleNet v1 and 25\% faster than Xception. On ARM, the speeds of ShuffleNet v1, Xception and ShuffleNet v2 are comparable; however, MobileNet v2 is much slower, especially on smaller FLOPs. We believe this is because MobileNet v2 has higher MAC (see \textbf{G1} and \textbf{G4} in Sec.~\ref{sec:guidelines}), which is significant on mobile devices.

Compared with \emph{MobileNet v1} \cite{howard2017mobilenets}, \emph{IGCV2} \cite{xie2018igcv}, and \emph{IGCV3} \cite{sun2018igcv3}, we have two observations. First, although the accuracy of MobileNet v1 is not as good, its speed on GPU is faster than all the counterparts, including ShuffleNet v2. We believe this is because its structure satisfies most of proposed guidelines (e.g. for \textbf{G3}, the fragments of MobileNet v1 are even fewer than ShuffleNet v2). Second, IGCV2 and IGCV3 are slow. This is due to usage of too many convolution groups (4 or 8 in \cite{xie2018igcv,sun2018igcv3}). Both observations are consistent with our proposed guidelines.

Recently, automatic model search~\cite{zoph2017learning,liu2017progressive,real2018regularized,xie2017genetic,real2017large,zoph2016neural} has become a promising trend for CNN architecture design. The bottom section in Table~\ref{tbl:cls_comp} evaluates some auto-generated models. We find that their speeds are relatively slow. We believe this is mainly due to the usage of too many fragments (see \textbf{G3}). Nevertheless, this research direction is still promising. Better models may be obtained, for example, if model search algorithms are combined with our proposed guidelines, and the direct metric (speed) is evaluated on the target platform.

Finally, Figure~\ref{fig:speed}(a)(b) summarizes the results of accuracy vs. speed, the direct metric. We conclude that ShuffeNet v2 is best on both GPU and ARM. 

\paragraph{Compatibility with other methods.} ShuffeNet v2 can be combined with other techniques to further advance the performance. When equipped with \emph{Squeeze-and-excitation} (SE) module~\cite{hu2017squeeze}, the classification accuracy of ShuffleNet v2 is improved by 0.5\% at the cost of certain loss in speed. The block structure is illustrated in Appendix Figure 2(b). Results are shown in Table~\ref{tbl:cls_comp} (bottom section).



\paragraph{Generalization to Large Models. } Although our main ablation is performed for light weight scenarios, ShuffleNet v2 can be used for large models (e.g, FLOPs $\ge$ 2G). Table~\ref{tbl:large_cls} compares a 50-layer ShuffleNet v2 (details in Appendix) with the counterpart of ShuffleNet v1 \cite{zhang2017shufflenet} and \emph{ResNet-50} \cite{he2016deep}. ShuffleNet v2 still outperforms ShuffleNet v1 at 2.3GFLOPs and surpasses ResNet-50 with 40\% fewer FLOPs.  

For very deep ShuffleNet v2 (e.g. over 100 layers), for the training to converge faster, we slightly modify the basic ShuffleNet v2 unit by adding a residual path (details in Appendix). Table~\ref{tbl:large_cls} presents a ShuffleNet v2 model of 164 layers equipped with \emph{SE} \cite{hu2017squeeze} components (details in Appendix). It obtains superior accuracy over the previous state-of-the-art models \cite{hu2017squeeze} with much fewer FLOPs.

\paragraph{Object Detection} To evaluate the generalization ability, we also tested COCO object detection \cite{lin2014microsoft} task. We use the state-of-the-art light-weight detector -- \emph{Light-Head RCNN} \cite{li2017light} -- as our framework and follow the same training and test protocols. Only backbone networks are replaced with ours. Models are pretrained on ImageNet and then finetuned on detection task. For training we use \emph{train+val} set in COCO except for 5000 images from \emph{minival} set, and use the \emph{minival} set to test. The accuracy metric is COCO standard \emph{mmAP}, i.e. the averaged mAPs at the box IoU thresholds from 0.5 to 0.95.

ShuffleNet v2 is compared with other three light-weight models: \emph{Xception} \cite{chollet2016xception,li2017light}, \emph{ShuffleNet v1} \cite{zhang2017shufflenet} and \emph{MobileNet v2} \cite{sandler2018inverted} on four levels of complexities. Results in Table~\ref{tbl:det} show that ShuffleNet v2 performs the best. 

Compared the detection result (Table~\ref{tbl:det}) with classification result (Table~\ref{tbl:cls_comp}), it is interesting that, on classification the accuracy rank is ShuffleNet v2 $\ge$ MobileNet v2 $>$ ShuffeNet v1 $>$ Xception, while on detection the rank becomes ShuffleNet v2 $>$ Xception $\ge$ ShuffleNet v1 $\ge$ MobileNet v2. This reveals that Xception is good on detection task. This is probably due to the larger receptive field of Xception building blocks than the other counterparts (7 \emph{vs.} 3). Inspired by this, we also enlarge the receptive field of ShuffleNet v2 by introducing an additional $3\times 3$ depthwise convolution before the first pointwise convolution in each building block. This variant is denoted as \emph{ShuffleNet v2*}. With only a few additional FLOPs, it further improves accuracy.

We also benchmark the runtime time on GPU. For fair comparison the batch size is set to 4 to ensure full GPU utilization. Due to the overheads of data copying (the resolution is as high as $800\times 1200$) and other detection-specific operations (like \emph{PSRoI Pooling} \cite{li2017light}), the speed gap between different models is smaller than that of classification. Still, ShuffleNet v2 outperforms others, e.g. around 40\% faster than ShuffleNet v1 and 16\% faster than MobileNet v2. 

Furthermore, the variant ShuffleNet v2* has best accuracy and is still faster than other methods. This motivates a practical question: how to increase the size of receptive field? This is critical for object detection in high-resolution images \cite{peng2017large}. We will study the topic in the future.

\begin{table}[t]
\centering
\begin{tabular}{lcc}
\hline
Model & FLOPs & Top-1 err. (\%) \\ 
\hline
ShuffleNet v2-50 (ours) & \textbf{2.3G}  & \textbf{22.8} \\
ShuffleNet v1-50 \cite{zhang2017shufflenet} (our impl.) & \textbf{2.3G}  & 25.2 \\
ResNet-50 \cite{he2016deep} & 3.8G  & 24.0 \\ 
\hline
SE-ShuffleNet v2-164 (ours, with residual) & \textbf{12.7G} & \textbf{18.56} \\ 
SENet \cite{hu2017squeeze} & 20.7G & 18.68 \\ 
\hline
\end{tabular}
\caption{Results of large models. See text for details.}
\label{tbl:large_cls}
\end{table}

\begin{table}[h]
\centering
\begin{tabular}{l|cccc|cccc}
\hline
Model                 & \multicolumn{4}{c|}{mmAP(\%)}                                          & \multicolumn{4}{c}{\begin{tabular}[c]{@{}c@{}}GPU Speed\\ (Images/sec.)\end{tabular}} \\ \hline
FLOPs                 & 40M             & 140M            & 300M            & 500M            & 40M                  & 140M                & 300M                & 500M               \\ \hline
Xception              & 21.9            & 29.0            & 31.3            & 32.9            & 178                  & 131                 & 101                 & 83                 \\
ShuffleNet v1         & 20.9            & 27.0            & 29.9            & 32.9            & 152                  & 85                  & 76                  & 60                 \\
MobileNet v2          & 20.7            & 24.4            & 30.0            & 30.6            & 146                  & 111                 & 94                  & 72                 \\
ShuffleNet v2 (ours)  & 22.5            & 29.0            & 31.8            & 33.3            & \textbf{188}       & \textbf{146}      & \textbf{109}      & \textbf{87}      \\
ShuffleNet v2* (ours) & \textbf{23.7} & \textbf{29.6} & \textbf{32.2} & \textbf{34.2} & 183                  & 138                 & 105                 & 83                \\ \hline
\end{tabular}

\caption{Performance on COCO object detection. The input image size is $800\times 1200$. \emph{FLOPs} row lists the complexity levels at $224\times 224$ input size. For GPU speed evaluation, the batch size is 4. We do not test ARM because the \emph{PSRoI Pooling} operation needed in~\cite{li2017light} is unavailable on ARM currently. }
\label{tbl:det}
\end{table}

\section{Conclusion}

\begin{table}[t]
\centering
\begin{tabular}{lcccc}
\hline
Model & \begin{tabular}[c]{@{}c@{}}Complexity\\  (MFLOPs)\end{tabular} & \begin{tabular}[c]{@{}c@{}}Top-1 \\ err. (\%)\end{tabular} & \begin{tabular}[c]{@{}c@{}}GPU Speed\\ (Batches/sec.)\end{tabular} & \begin{tabular}[c]{@{}c@{}}ARM Speed\\ (Images/sec.)\end{tabular} \\ 
\hline
ShuffleNet v2 0.5$\times$ (ours) &  \underline{41} & \underline{\textbf{39.7}} & \underline{417} & \underline{\textbf{57.0}}\\
0.25 MobileNet v1 \cite{howard2017mobilenets}  & 41 & 49.4 & \textbf{502} & 36.4 \\
0.4 MobileNet v2 \cite{sandler2018inverted} (our impl.)\textsuperscript{*} & 43 & 43.4 & 333 & 33.2 \\
0.15 MobileNet v2 \cite{sandler2018inverted} (our impl.) & 39 & 55.1 & 351 & 33.6 \\
 ShuffleNet v1 0.5$\times$ (g=3) \cite{zhang2017shufflenet} & 38 & 43.2 & 347 & 56.8 \\
 DenseNet 0.5$\times$ \cite{huang2017densely}  (our impl.) & 42 & 58.6 & 366 & 39.7 \\
 Xception 0.5$\times$ \cite{chollet2016xception}  (our impl.) & 40 & 44.9 & 384 & 52.9 \\ 
 IGCV2-0.25 \cite{xie2018igcv} & 46 & 45.1 & 183 & 31.5 \\
 \hline 
 ShuffleNet v2 1$\times$ (ours) & \underline{146} & \underline{\textbf{30.6}} & \underline{341} & \underline{\textbf{24.4}} \\
0.5 MobileNet v1 \cite{howard2017mobilenets} & 149 & 36.3 & \textbf{382} & 16.5 \\
 0.75 MobileNet v2 \cite{sandler2018inverted} (our impl.)\textsuperscript{**} & 145 & 32.1 & 235 & 15.9 \\
 0.6 MobileNet v2 \cite{sandler2018inverted} (our impl.) & 141 & 33.3 & 249 & 14.9 \\
 ShuffleNet v1 1$\times$ (g=3)  \cite{zhang2017shufflenet}  & 140 & 32.6 & 213 & 21.8 \\
 DenseNet 1$\times$   \cite{huang2017densely}   (our impl.) & 142 & 45.2 & 279 & 15.8 \\
 Xception 1$\times$     \cite{chollet2016xception}   (our impl.) & 145 & 34.1 & 278 & 19.5 \\ 
 IGCV2-0.5 \cite{xie2018igcv} & 156 & 34.5 & 132 & 15.5 \\
 IGCV3-D (0.7) \cite{sun2018igcv3} & 210 & 31.5 & 143 & 11.7 \\
 \hline
 ShuffleNet v2 1.5$\times$ (ours)    & \underline{299} & \underline{\textbf{27.4}} & \underline{255} & \underline{\textbf{11.8}} \\
 0.75 MobileNet v1 \cite{howard2017mobilenets} & 325 & 31.6 & \textbf{314} & 10.6 \\
1.0 MobileNet v2    \cite{sandler2018inverted}    & 300 & 28.0 & 180 & 8.9 \\
 1.0 MobileNet v2 \cite{sandler2018inverted} (our impl.) & 301 & 28.3 & 180 & 8.9 \\
 ShuffleNet v1 1.5$\times$ (g=3)  \cite{zhang2017shufflenet}& 292 & 28.5 & 164 & 10.3 \\
 DenseNet 1.5$\times$   \cite{huang2017densely}  (our impl.)     & 295 & 39.9 & 274 & 9.7\\
 CondenseNet (G=C=8) \cite{huang2017condensenet} & 274 & 29.0 & - & -\\
 Xception 1.5$\times$  \cite{chollet2016xception}   (our impl.) & 305 & 29.4 & 219 & 10.5 \\ 
 IGCV3-D \cite{sun2018igcv3} & 318 & 27.8 & 102 & 6.3 \\
 \hline
ShuffleNet v2 2$\times$ (ours) & \underline{591} & \underline{\textbf{25.1}} & \underline{217} & \underline{\textbf{6.7}} \\
1.0 MobileNet v1 \cite{howard2017mobilenets}    & 569 & 29.4 & \textbf{247} & 6.5 \\
1.4 MobileNet v2  \cite{sandler2018inverted}     & 585 & 25.3 & 137 & 5.4 \\
 1.4 MobileNet v2 \cite{sandler2018inverted} (our impl.) & 587 & 26.7 & 137 & 5.4 \\
 ShuffleNet v1 2$\times$ (g=3)   \cite{zhang2017shufflenet} & 524 & 26.3 & 133 & 6.4 \\
 DenseNet 2$\times$   \cite{huang2017densely}  (our impl.)    & 519 & 34.6 & 197 & 6.1 \\
CondenseNet (G=C=4)    \cite{huang2017condensenet}    & 529 & 26.2 & - & - \\
Xception 2$\times$  \cite{chollet2016xception}   (our impl.) & 525 & 27.6 & 174 & \textbf{6.7} \\ 
IGCV2-1.0 \cite{xie2018igcv} & 564 & 29.3 & 81 & 4.9 \\
IGCV3-D (1.4) \cite{sun2018igcv3} & 610 & 25.5 & 82 & 4.5 \\
\hline\hline
ShuffleNet v2 2x (ours, with \emph{SE} \cite{hu2017squeeze}) & \underline{597}  & \underline{\textbf{24.6}} & \underline{\textbf{161}} & \underline{\textbf{5.6}} \\
NASNet-A \cite{zoph2017learning} ( 4 $@$ 1056, our impl.) & 564  & 26.0 & 130 & 4.6 \\
PNASNet-5 \cite{liu2017progressive} (our impl.)  & 588  & 25.8 & 115 & 4.1 \\ 
\hline 
\end{tabular}
\caption{Comparison of several network architectures over classification error (on validation set, single center crop) and speed, on two platforms and four levels of computation complexity. Results are grouped by complexity levels for better comparison. The batch size is 8 for GPU and 1 for ARM. The image size is $224\times 224$ except: [*] $160\times 160$ and [**] $192\times 192$. We do not provide speed measurements for \emph{CondenseNets} \cite{huang2017condensenet} due to lack of efficient implementation currently.}

\label{tbl:cls_comp}
\end{table}

We propose that network architecture design should consider the direct metric such as speed, instead of the indirect metric like FLOPs. We present practical guidelines and a novel architecture, ShuffleNet v2. Comprehensive experiments verify the effectiveness of our new model. We hope this work could inspire future work of network architecture design that is platform aware and more practical.

\paragraph{\textbf{Acknowledgements}}

Thanks Yichen Wei for his help on paper writing. This research is partially supported by National Natural Science Foundation of China (Grant No. 61773229).

\clearpage
\bibliographystyle{splncs}
\bibliography{egbib}

\begin{thebibliography}{10}

\bibitem{krizhevsky2012imagenet}
Krizhevsky, A., Sutskever, I., Hinton, G.E.:
\newblock Imagenet classification with deep convolutional neural networks.
\newblock In: Advances in neural information processing systems. (2012)
  1097--1105

\bibitem{simonyan2014very}
Simonyan, K., Zisserman, A.:
\newblock Very deep convolutional networks for large-scale image recognition.
\newblock arXiv preprint arXiv:1409.1556 (2014)

\bibitem{szegedy2015going}
Szegedy, C., Liu, W., Jia, Y., Sermanet, P., Reed, S., Anguelov, D., Erhan, D.,
  Vanhoucke, V., Rabinovich, A.,  et~al.:
\newblock Going deeper with convolutions, Cvpr (2015)

\bibitem{he2016deep}
He, K., Zhang, X., Ren, S., Sun, J.:
\newblock Deep residual learning for image recognition.
\newblock In: Proceedings of the IEEE conference on computer vision and pattern
  recognition. (2016)  770--778

\bibitem{he2016identity}
He, K., Zhang, X., Ren, S., Sun, J.:
\newblock Identity mappings in deep residual networks.
\newblock In: European Conference on Computer Vision, Springer (2016)  630--645

\bibitem{huang2017densely}
Huang, G., Liu, Z., Weinberger, K.Q., van~der Maaten, L.:
\newblock Densely connected convolutional networks.
\newblock In: Proceedings of the IEEE conference on computer vision and pattern
  recognition. Volume~1. (2017) ~3

\bibitem{xie2017aggregated}
Xie, S., Girshick, R., Doll{\'a}r, P., Tu, Z., He, K.:
\newblock Aggregated residual transformations for deep neural networks.
\newblock In: Computer Vision and Pattern Recognition (CVPR), 2017 IEEE
  Conference on, IEEE (2017)  5987--5995

\bibitem{hu2017squeeze}
Hu, J., Shen, L., Sun, G.:
\newblock Squeeze-and-excitation networks.
\newblock arXiv preprint arXiv:1709.01507 (2017)

\bibitem{zoph2017learning}
Zoph, B., Vasudevan, V., Shlens, J., Le, Q.V.:
\newblock Learning transferable architectures for scalable image recognition.
\newblock arXiv preprint arXiv:1707.07012 (2017)

\bibitem{liu2017progressive}
Liu, C., Zoph, B., Shlens, J., Hua, W., Li, L.J., Fei-Fei, L., Yuille, A.,
  Huang, J., Murphy, K.:
\newblock Progressive neural architecture search.
\newblock arXiv preprint arXiv:1712.00559 (2017)

\bibitem{real2018regularized}
Real, E., Aggarwal, A., Huang, Y., Le, Q.V.:
\newblock Regularized evolution for image classifier architecture search.
\newblock arXiv preprint arXiv:1802.01548 (2018)

\bibitem{chollet2016xception}
Chollet, F.:
\newblock Xception: Deep learning with depthwise separable convolutions.
\newblock arXiv preprint (2016)

\bibitem{howard2017mobilenets}
Howard, A.G., Zhu, M., Chen, B., Kalenichenko, D., Wang, W., Weyand, T.,
  Andreetto, M., Adam, H.:
\newblock Mobilenets: Efficient convolutional neural networks for mobile vision
  applications.
\newblock arXiv preprint arXiv:1704.04861 (2017)

\bibitem{sandler2018inverted}
Sandler, M., Howard, A., Zhu, M., Zhmoginov, A., Chen, L.C.:
\newblock Inverted residuals and linear bottlenecks: Mobile networks for
  classification, detection and segmentation.
\newblock arXiv preprint arXiv:1801.04381 (2018)

\bibitem{zhang2017shufflenet}
Zhang, X., Zhou, X., Lin, M., Sun, J.:
\newblock Shufflenet: An extremely efficient convolutional neural network for
  mobile devices.
\newblock arXiv preprint arXiv:1707.01083 (2017)

\bibitem{huang2017condensenet}
Huang, G., Liu, S., van~der Maaten, L., Weinberger, K.Q.:
\newblock Condensenet: An efficient densenet using learned group convolutions.
\newblock arXiv preprint arXiv:1711.09224 (2017)

\bibitem{liu2017learning}
Liu, Z., Li, J., Shen, Z., Huang, G., Yan, S., Zhang, C.:
\newblock Learning efficient convolutional networks through network slimming.
\newblock In: 2017 IEEE International Conference on Computer Vision (ICCV),
  IEEE (2017)  2755--2763

\bibitem{wen2016learning}
Wen, W., Wu, C., Wang, Y., Chen, Y., Li, H.:
\newblock Learning structured sparsity in deep neural networks.
\newblock In: Advances in Neural Information Processing Systems. (2016)
  2074--2082

\bibitem{he2017channel}
He, Y., Zhang, X., Sun, J.:
\newblock Channel pruning for accelerating very deep neural networks.
\newblock In: International Conference on Computer Vision (ICCV). Volume~2.
  (2017) ~6

\bibitem{jaderberg2014speeding}
Jaderberg, M., Vedaldi, A., Zisserman, A.:
\newblock Speeding up convolutional neural networks with low rank expansions.
\newblock arXiv preprint arXiv:1405.3866 (2014)

\bibitem{zhang2015efficient}
Zhang, X., Zou, J., Ming, X., He, K., Sun, J.:
\newblock Efficient and accurate approximations of nonlinear convolutional
  networks.
\newblock In: Proceedings of the IEEE Conference on Computer Vision and Pattern
  Recognition. (2015)  1984--1992

\bibitem{zhang2016accelerating}
Zhang, X., Zou, J., He, K., Sun, J.:
\newblock Accelerating very deep convolutional networks for classification and
  detection.
\newblock IEEE transactions on pattern analysis and machine intelligence
  \textbf{38}(10) (2016)  1943--1955

\bibitem{chetlur2014cudnn}
Chetlur, S., Woolley, C., Vandermersch, P., Cohen, J., Tran, J., Catanzaro, B.,
  Shelhamer, E.:
\newblock cudnn: Efficient primitives for deep learning.
\newblock arXiv preprint arXiv:1410.0759 (2014)

\bibitem{das2016distributed}
Das, D., Avancha, S., Mudigere, D., Vaidynathan, K., Sridharan, S., Kalamkar,
  D., Kaul, B., Dubey, P.:
\newblock Distributed deep learning using synchronous stochastic gradient
  descent.
\newblock arXiv preprint arXiv:1602.06709 (2016)

\bibitem{ioannou2016deep}
Ioannou, Y., Robertson, D., Cipolla, R., Criminisi, A.:
\newblock Deep roots: Improving cnn efficiency with hierarchical filter groups.
\newblock arXiv preprint arXiv:1605.06489 (2016)

\bibitem{zhang2017interleaved}
Zhang, T., Qi, G.J., Xiao, B., Wang, J.:
\newblock Interleaved group convolutions for deep neural networks.
\newblock In: International Conference on Computer Vision. (2017)

\bibitem{xie2018igcv}
Xie, G., Wang, J., Zhang, T., Lai, J., Hong, R., Qi, G.J.:
\newblock Igcv $2 $: Interleaved structured sparse convolutional neural
  networks.
\newblock arXiv preprint arXiv:1804.06202 (2018)

\bibitem{sun2018igcv3}
Sun, K., Li, M., Liu, D., Wang, J.:
\newblock Igcv3: Interleaved low-rank group convolutions for efficient deep
  neural networks.
\newblock arXiv preprint arXiv:1806.00178 (2018)

\bibitem{szegedy2017inception}
Szegedy, C., Ioffe, S., Vanhoucke, V., Alemi, A.A.:
\newblock Inception-v4, inception-resnet and the impact of residual connections
  on learning.
\newblock In: AAAI. Volume~4. (2017) ~12

\bibitem{szegedy2016rethinking}
Szegedy, C., Vanhoucke, V., Ioffe, S., Shlens, J., Wojna, Z.:
\newblock Rethinking the inception architecture for computer vision.
\newblock In: Proceedings of the IEEE Conference on Computer Vision and Pattern
  Recognition. (2016)  2818--2826

\bibitem{ioffe2015batch}
Ioffe, S., Szegedy, C.:
\newblock Batch normalization: Accelerating deep network training by reducing
  internal covariate shift.
\newblock In: International conference on machine learning. (2015)  448--456

\bibitem{deng2009imagenet}
Deng, J., Dong, W., Socher, R., Li, L.J., Li, K., Fei-Fei, L.:
\newblock Imagenet: A large-scale hierarchical image database.
\newblock In: Computer Vision and Pattern Recognition, 2009. CVPR 2009. IEEE
  Conference on, IEEE (2009)  248--255

\bibitem{russakovsky2015imagenet}
Russakovsky, O., Deng, J., Su, H., Krause, J., Satheesh, S., Ma, S., Huang, Z.,
  Karpathy, A., Khosla, A., Bernstein, M.,  et~al.:
\newblock Imagenet large scale visual recognition challenge.
\newblock International Journal of Computer Vision \textbf{115}(3) (2015)
  211--252

\bibitem{li2017light}
Li, Z., Peng, C., Yu, G., Zhang, X., Deng, Y., Sun, J.:
\newblock Light-head r-cnn: In defense of two-stage object detector.
\newblock arXiv preprint arXiv:1711.07264 (2017)

\bibitem{xie2017genetic}
Xie, L., Yuille, A.:
\newblock Genetic cnn.
\newblock arXiv preprint arXiv:1703.01513 (2017)

\bibitem{real2017large}
Real, E., Moore, S., Selle, A., Saxena, S., Suematsu, Y.L., Le, Q., Kurakin,
  A.:
\newblock Large-scale evolution of image classifiers.
\newblock arXiv preprint arXiv:1703.01041 (2017)

\bibitem{zoph2016neural}
Zoph, B., Le, Q.V.:
\newblock Neural architecture search with reinforcement learning.
\newblock arXiv preprint arXiv:1611.01578 (2016)

\bibitem{lin2014microsoft}
Lin, T.Y., Maire, M., Belongie, S., Hays, J., Perona, P., Ramanan, D.,
  Doll{\'a}r, P., Zitnick, C.L.:
\newblock Microsoft coco: Common objects in context.
\newblock In: European conference on computer vision, Springer (2014)  740--755

\bibitem{peng2017large}
Peng, C., Zhang, X., Yu, G., Luo, G., Sun, J.:
\newblock Large kernel matters--improve semantic segmentation by global
  convolutional network.
\newblock arXiv preprint arXiv:1703.02719 (2017)

\end{thebibliography}
\clearpage

\section*{Appendix}

\setcounter{table}{0}
\captionsetup[table]{name= Appendix Table}

\setcounter{figure}{0}
\captionsetup[figure]{name= Appendix Fig.}

\begin{figure}[h]
	\centering
	\includegraphics[height=3.3cm]{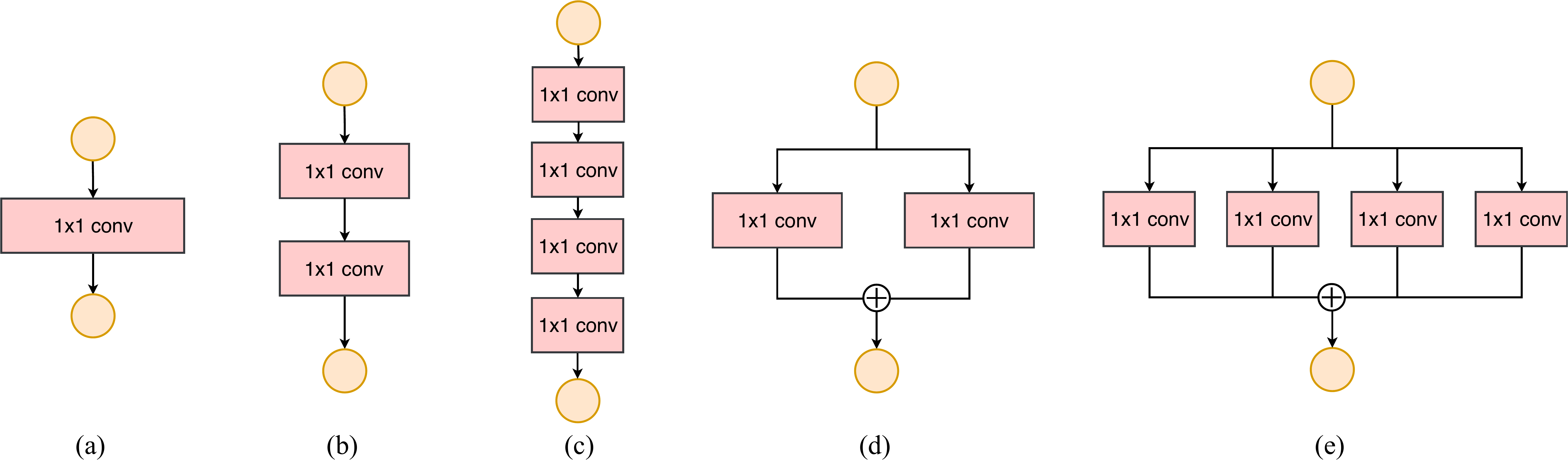}
	\caption{Building blocks used in experiments for guideline 3. (a) \emph{1-fragment}. (b) \emph{2-fragment-series}. (c) \emph{4-fragment-series}. (d) \emph{2-fragment-parallel}. (e) \emph{4-fragment-parallel}. } 
	\label{fig:guideline_fragmentation}
\end{figure}


\begin{table}[h]
\centering
\begin{subtable}{\textwidth}
\centering
\begin{tabular}{l|l|cccc|cccc}
\hline
                          &           & \multicolumn{4}{c}{GPU (Batches/sec.)} & \multicolumn{4}{|c}{CPU (Images/sec.)} \\ \hline
  Input size         & FLOPs  & 40M     & 140M     & 300M    & 500M    & 40M     & 140M    & 300M    & 500M    \\ \hline
\multirow{4}{*}{320x320}  & ShuffleNet v2 & 315$\ast$     & 525      & 474     & 422     & 28.1    & 12.5    & 6.1     & 3.4     \\ \cline{2-10} 
                          & ShuffleNet v1 & 236$\ast$     & 414      & 344     & 275     & 27.2    & 11.4    & 5.1     & 3.1     \\ \cline{2-10} 
                          & MobileNet v2  & 187$\ast$     & 460      & 389     & 335     & 11.4    & 6.4     & 4.6     & 2.7     \\ \cline{2-10} 
                          & Xception  & 279$\ast$     & 463      & 408     & 350     & 31.1    & 10.1    & 5.6     & 3.5     \\ \hline
\multirow{4}{*}{640x480}  & ShuffleNet v2 & 424     & 394      & 297     & 250     & 9.3     & 4.0     & 1.9     & 1.1     \\ \cline{2-10} 
                          & ShuffleNet v1 & 396     & 269      & 198     & 156     & 8.0     & 3.7     & 1.6     & 1.0     \\ \cline{2-10} 
                          & MobileNet v2  & 338     & 248      & 208     & 165     & 3.8     & 2.0     & 1.4     & 0.8     \\ \cline{2-10} 
                          & Xception  & 399     & 326      & 244     & 209     & 9.6     & 3.2     & 1.7     & 1.1     \\ \hline
\multirow{4}{*}{1080x720} & ShuffleNet v2 & 248     & 197      & 141     & 115     & 3.5     & 1.5     & 0.7     & 0.4     \\ \cline{2-10} 
                          & ShuffleNet v1 & 203     & 131      & 96      & 77      & 2.9     & 1.4     & 0.4     & 0.3     \\ \cline{2-10} 
                          & MobileNet v2  & 159     & 117      & 99      & 78      & 1.4     & 0.7     & 0.3     & 0.3     \\ \cline{2-10} 
                          & Xception  & 232     & 160      & 124     & 106     & 3.6     & 1.2     & 0.5     & 0.4     \\ \hline
\end{tabular}
\caption{Comparison of actual speeds (whole architecture).}
\label{tabel:wholenet}
\vspace{10mm}
\end{subtable} \\
\begin{subtable}{\textwidth}
\centering
\begin{tabular}{c|c|ccc|ccc}
\hline
                       &           & \multicolumn{3}{c}{GPU (Batches/sec.)} & \multicolumn{3}{|c}{CPU (Images/sec.)} \\ \hline
   Input size  & Channel (c) for ShuffleNet v2 & c=64        & c=128       & c=256      & c=64        & c=128      & c=256      \\ \hline
\multirow{4}{*}{56x56} & ShuffleNet v2 & 216      & 142      & 81      & 34.8        & 12.3       & 3.9        \\ \cline{2-8} 
                       & ShuffleNet v1 & 127      & 73       & 45      & 24.3        & 9.4        & 3.0        \\ \cline{2-8} 
                       & MobileNet v2  & 89       & 125      & 69      & 25.8        & 10.0       & 3.0        \\ \cline{2-8} 
                       & Xception  & 185      & 52       & 68      & 27.0        & 9.7        & 3.1        \\ \hline
\multirow{4}{*}{28x28} & ShuffleNet v2 & 407      & 313      & 237     & 174.5       & 53.4       & 16.6       \\ \cline{2-8} 
                       & ShuffleNet v1 & 298      & 222      & 60      & 139.7       & 43.9       & 13.2       \\ \cline{2-8} 
                       & MobileNet v2  & 381      & 286      & 189     & 118.3       & 46.2       & 13.3       \\ \cline{2-8} 
                       & Xception  & 254      & 238      & 169     & 117.0       & 45.8       & 14.0       \\ \hline
\end{tabular}
\caption{Comparison of actual speeds (units).}
\label{table:blocks}
\vspace{10mm}
\end{subtable}
\caption{Table (a) compares the speed of each network (whole architecture). Table (b) compares the speed of each network's unit, we stack 10 network units of each network; the value of $c$ means the number of channels for ShuffleNet v2, we adjust the number of channels to keep the FLOPs unchanged for other network units. Please refer to Section 4 for details. $\left [  \ast   \right ]$ For the models of 40M FLOPs with input size of $320 \times 320$, the $batchsize$ is set to 8 to ensure the GPU utilization rate, and we set $batchsize=1$ otherwise. }
\end{table}

\begin{table}[]
\centering
\begin{tabular}{c|c|c|c|c|c}
\hline
layer                & \begin{tabular}[c]{@{}c@{}}output \\ size\end{tabular} & \begin{tabular}[c]{@{}c@{}}ShuffleNet v1-50 \\ (group=3) \end{tabular}                                                             & ShuffleNet v2-50                                                            & Resnet-50                                                                   & \begin{tabular}[c]{@{}c@{}}SE-ShuffleNet\\ v2-164 \end{tabular}                                                              \\ \hline
conv1\_x                  & 112$\times$112                & 3$\times$3, 64, stride 2                                                             & 3$\times$3, 64, stride 2                                                            & 7$\times$7, 64, stride 2                                                           & \begin{math}  \begin{tabular}[c]{@{}c@{}}3$\times$3, 64, stride 2\\ 3$\times$3, 64\\ 3$\times$3, 128\end{tabular}  \end{math} \\ \hline
\multirow{2}{*}{conv2\_x} & \multirow{2}{*}{56$\times$56} & \multicolumn{4}{c}{3$\times$3 max pool, stride 2}                                                                                                                                                                                                                                                                                  \\ \cline{3-6} 
                          &                        & \begin{math} \left[ \begin{tabular}[c]{@{}c@{}}1$\times$1, 360\\ 3$\times$3, 360\\ 1$\times$1, 360  \end{tabular} \right ] \end{math}$\times$3  & \begin{math} \left[ \begin{tabular}[c]{@{}c@{}}1$\times$1, 244\\ 3$\times$3, 244\\ 1$\times$1, 244  \end{tabular}  \right ] \end{math}$\times$3  & \begin{math} \left[ \begin{tabular}[c]{@{}c@{}}1$\times$1, 64\\ 3$\times$3, 64\\ 1$\times$1, 256 \end{tabular}  \right ] \end{math}$\times$3   & \begin{math} \left[ \begin{tabular}[c]{@{}c@{}}1$\times$1, 340\\  3$\times$3, 340\\ 1$\times$1, 340  \end{tabular} \right ] \end{math}$\times$10 \\ \hline
conv3\_x                  & 28$\times$28                  & \begin{math} \left[ \begin{tabular}[c]{@{}c@{}}1$\times$1, 720\\ 3$\times$3, 720\\ 1$\times$1, 720 \end{tabular} \right ] \end{math}$\times$4 & \begin{math} \left[ \begin{tabular}[c]{@{}c@{}}1$\times$1, 488\\ 3$\times$3, 488\\ 1$\times$1, 488 \end{tabular} \right ] \end{math}$\times$4 &  \begin{math} \left[  \begin{tabular}[c]{@{}c@{}}1$\times$1, 128\\  3$\times$3, 128\\ 1$\times$1, 512 \end{tabular}  \right ] \end{math}$\times$4 & \begin{math} \left[ \begin{tabular}[c]{@{}c@{}}1$\times$1, 680\\  3$\times$3, 680\\  1$\times$1, 680 \end{tabular} \right ] \end{math}$\times$10 \\ \hline
conv4\_x                  & 14$\times$14                  & \begin{math} \left[ \begin{tabular}[c]{@{}c@{}}1$\times$1, 1440\\ 3$\times$3, 1440\\ 1$\times$1, 1440 \end{tabular} \right ] \end{math}$\times$6 & \begin{math} \left[ \begin{tabular}[c]{@{}c@{}}1$\times$1, 976\\ 3$\times$3, 976\\ 1$\times$1, 976 \end{tabular} \right ] \end{math}$\times$6 & \begin{math} \left[ \begin{tabular}[c]{@{}c@{}}1$\times$1, 256\\  3$\times$3, 256\\ 1$\times$1, 1024 \end{tabular} \right ] \end{math}$\times$6 & \begin{math} \left[  \begin{tabular}[c]{@{}c@{}}1$\times$1, 1360\\  3$\times$3, 1360\\  1$\times$1, 1360 \end{tabular} \right ] \end{math}$\times$23 \\ \hline
conv5\_x                  & 7$\times$7                    & \begin{math} \left[ \begin{tabular}[c]{@{}c@{}}1$\times$1, 2880\\ 3$\times$3, 2880\\ 1$\times$1, 2880 \end{tabular} \right ] \end{math}$\times$3 & \begin{math} \left[ \begin{tabular}[c]{@{}c@{}}1$\times$1, 1952\\ 3$\times$3, 1952\\ 1$\times$1, 1952 \end{tabular} \right ] \end{math}$\times$3 & \begin{math} \left[ \begin{tabular}[c]{@{}c@{}}1$\times$1, 512\\ 3$\times$3, 512\\ 1$\times$1, 2048 \end{tabular} \right ] \end{math}$\times$3 & \begin{math} \left[ \begin{tabular}[c]{@{}c@{}}1$\times$1, 2720\\  3$\times$3, 2720\\  1$\times$1, 2720 \end{tabular} \right ] \end{math}$\times$10 \\ \hline
conv6                     & 7$\times$7                    & -                                                                            & 1$\times$1, 2048                                                                   & -                                                                           & 1$\times$1, 2048                                                                       \\ \hline
                          & 1$\times$1                    & \multicolumn{4}{c}{average pool, 1000-d fc, softmax}                                                                                                                                                                                                                                                                       \\ \hline
\multicolumn{2}{c|}{FLOPs}                          & 2.3G                                                                         & 2.3G                                                                        & 3.8G                                                                        & 12.7G                                                                           \\ \hline
\end{tabular}
\caption{Architectures for large models. Building blocks are shown in brackets, with the convolution kernel shapes and the numbers of blocks stacked. Downsampling is performed by $conv3\_1, conv4\_1,$ and $conv5\_1$ with a stride of 2. For ShuffleNet v1-50 and ResNet-50, the bottleneck ratio is set to 1:4. For SE-ShuffleNet v2-164, we add the SE modules right before the residual \emph{add-ReLUs} (details in Appendix Figure 2); we set the neural numbers in SE modules to the 1/2 of the channel numbers in the corresponding building blocks. See Section 4 for details.}
\label{tbl:structures}
\end{table}

\begin{figure}[h]
	\centering
	\includegraphics[height=6.9cm]{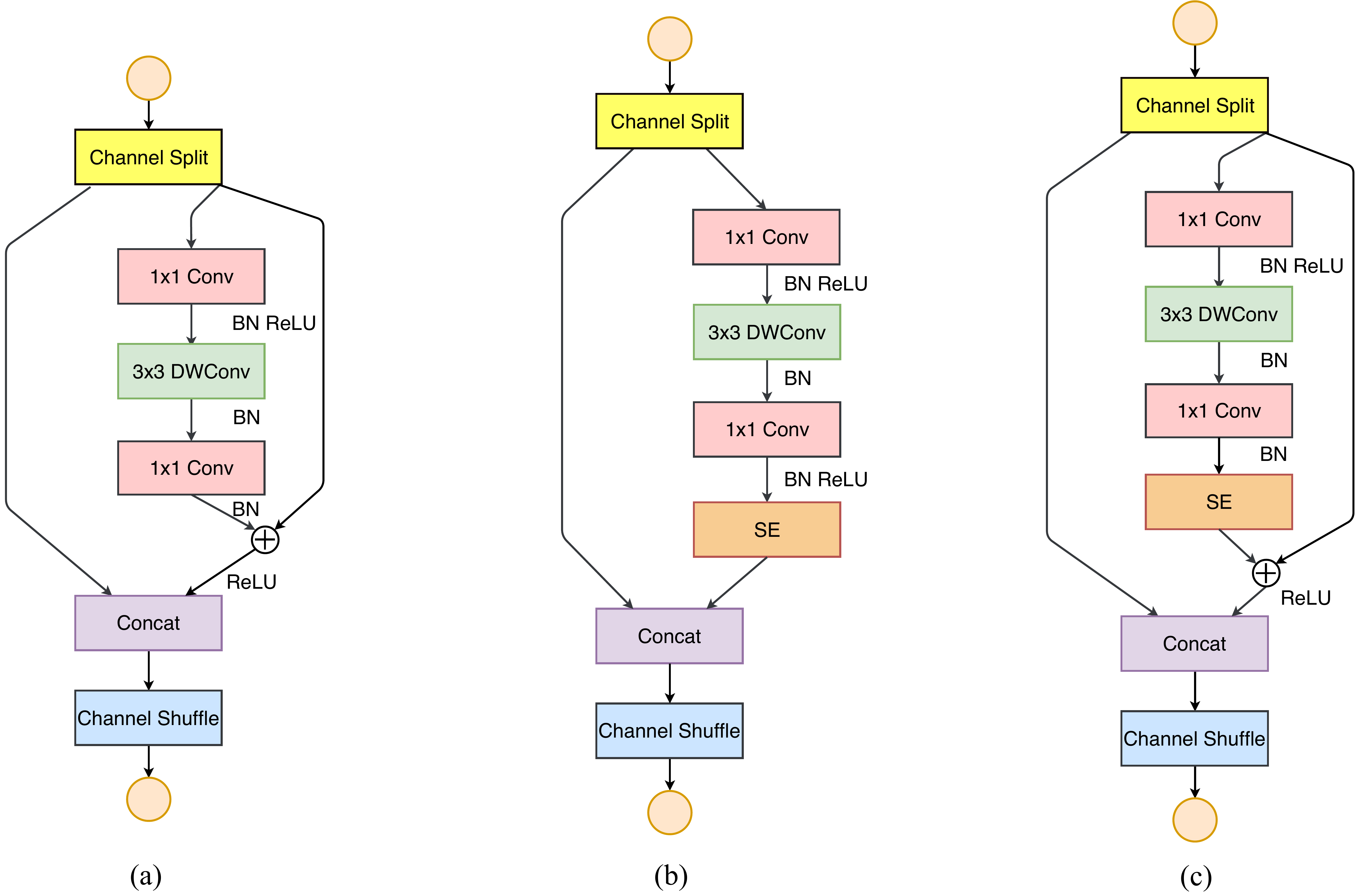}
	\caption{Building blocks of ShuffleNet v2 with SE/residual. (a) \emph{ShuffleNet v2 with residual}. (b) \emph{ShuffleNet v2 with SE}. (c) \emph{ShuffleNet v2 with SE and residual}. } 
	\label{fig:snet_deep}
\end{figure}

\clearpage

\end{document}